\documentclass[lettersize,journal]{IEEEtran}
\usepackage{amsmath,amsfonts}
\usepackage{algorithmic}
\usepackage{algorithm}
\usepackage{array}
\usepackage[caption=false,font=normalsize,labelfont=sf,textfont=sf]{subfig}
\usepackage{caption}
\usepackage{tabularx, booktabs}
\usepackage{textcomp}
\usepackage{stfloats}
\usepackage{url}
\usepackage{verbatim}
\usepackage{graphicx}
\usepackage{cite}
\usepackage[dvipsnames]{xcolor}

\usepackage{color,soul}
\setulcolor{red}
\sethlcolor{yellow}

\renewcommand\hl[1]{#1} 

\begin{document}
\bstctlcite{IEEEexample:BSTcontrol}

\title{An Efficient and Generalizable Transfer Learning Method for Weather Condition Detection on Ground Terminals}








\author{
Wenxuan Zhang~\IEEEmembership{}and
Peng Hu,~\IEEEmembership{Senior Member,~IEEE}
\thanks{Wenxuan Zhang is with Center for Center for Computational Mathematics, Faculty of Mathematics, University of Waterloo, Canada (e-mail: v39zhang@uwaterloo.ca)}
\thanks{Peng Hu is with Department of Electrical and Computer Engineering, University of Manitoba, Canada, and Faculty of Mathematics, University of Waterloo, Canada (e-mail: peng.hu@umanitoba.ca) (Corresponding author)}
\thanks{We acknowledge the support of the Natural Sciences and Engineering Research Council of Canada (NSERC), [funding reference number RGPIN-2022-03364].}
\thanks{}}


\markboth{Journal of \LaTeX\ Class Files,~Vol.~0, No.~0, 00~0000}%
{Shell \MakeLowercase{\textit{et al.}}: A Sample Article Using IEEEtran.cls for IEEE Journals}


\markboth{IEEE Transactions on Aerospace and Electronic Systems, DOI: 10.1109/TAES.2024.3496857}{}

\maketitle

\begin{abstract}
The increasing adoption of satellite Internet with low-Earth-orbit (LEO) satellites in mega-constellations allows ubiquitous connectivity to rural and remote areas. However, weather events have a significant impact on the performance and reliability of satellite Internet. Adverse weather events such as snow and rain can disturb the performance and operations of satellite Internet's essential ground terminal components, such as satellite antennas, significantly disrupting the space-ground link conditions between LEO satellites and ground stations. This challenge calls for not only region-based weather forecasts but also fine-grained detection capability on ground terminal components of fine-grained weather conditions. Such a capability can assist in fault diagnostics and mitigation for reliable satellite Internet, but its solutions are lacking, not to mention the effectiveness and generalization that are essential in real-world deployments. This paper discusses an efficient transfer learning (TL) method that can enable a ground component to locally detect representative weather-related conditions. The proposed method can detect snow, wet, and other conditions resulting from adverse and typical weather events and shows superior performance compared to the typical deep learning methods, such as YOLOv7, \hl{YOLOv9}, \hl{Faster R-CNN}, and R-YOLO. Our TL method also shows the advantage of being generalizable to various scenarios.
\end{abstract}

\begin{IEEEkeywords}
Transfer learning, ground terminal, satellite antenna, network reliability
\end{IEEEkeywords}

\section{Introduction}
\IEEEPARstart{B}{roadband} satellite Internet provided by low-Earth-orbit (LEO) satellites can enable ubiquitous connectivity for everyone on Earth. However, the reliable performance of such satellite Internet depends on the weather conditions, which can disturb the performance and operations of ground components that are essential for satellite Internet, satellite antennas used by emerging satellite ground stations and user terminals. They can then significantly disrupt the uplinks and downlinks that are essential for satellite Internet. Such an issue is evident in the recent studies \cite{Padmanabhan19, Laniewski2024WetLinksAL} where some weather conditions such as rain and snow are identified to have a significant impact on the LEO satellite network performance. For example, rain and snow weather events can attenuate radiofrequency (RF) signals in the Ka/Ku bands used on advanced LEO satellites for ground-space links, and cause increased failure rates \cite{Padmanabhan19}. General weather forecasts are not helpful for weather condition inference due to the complexity of the conditions on a ground component that can result from weather events. A computer vision based sensing solution shows great promise for detecting and recognizing the specific conditions of a ground component but how to design an efficient and generalizable method and apply it to the essential ground components, such as satellite antennas, has not been addressed in the literature.

In general, object detection based on convolutional neural networks (CNNs) for computer vision can be classified into one-stage and two-stage detectors. Two-stage detectors, such as R-CNN \cite{girshick2014rich}, prioritize detection accuracy by identifying regions of interest and then classifying and refining the bounding boxes. In contrast, one-stage detectors, such as You Only Look Once (YOLO) series \cite{wang2023yolov7} and Single Shot Detectors (SSD) \cite{10.1007/978-3-319-46448-0_2}, streamline this process by directly predicting bounding boxes and class confidences. One-stage detectors face challenges in detecting small objects due to their fixed bounding box scales and aspect ratios, but they offer faster training times and require fewer training images \cite{Leng2019An, 10.1007/978-3-319-46448-0_2}. R-YOLO, proposed by Wang \textit{et al.} \cite{9989413}, has specifically enhanced YOLO for adverse weather. Furthermore, most current research on object detection models aims to enhance their accuracy, typically requiring large datasets. However, irrelevant features can significantly impact the performance of these models in specific applications. Additionally, there is often a lack of integration of human knowledge in the model development process. Incorporating this knowledge could improve accuracy for particular problems without increasing the model's complexity, which is especially beneficial in scenarios with limited data availability.

Transfer learning (TL) is a machine learning (ML) technique that adapts knowledge learned from one task to another related task. The study \cite{Xu2020Transfer} highlights that the TL method is particularly advantageous when labelled data is scarce or specific conditions are underrepresented in domains. 

To the best of our knowledge, our work first provides a generalizable vision-based solution to weather condition detection for ground terminals. By applying TL, we address the challenge of data scarcity by adapting a model to classify the conditions of objects across varied weather conditions accurately. Feature-based TL specifically focuses on manipulating the data features extracted by the pre-trained model to better align with the specifics of the target application.

In this paper, we focus on designing an effective TL method for detecting weather-related conditions on satellite antennas. The You Only Look At CoefficienTs (YOLACT) framework is used to apply model transfer for segmentation and feature isolation. We will simulate an environment with limited data and address the challenge of data scarcity by adapting our model to classify the conditions of satellite antennas in various weather conditions. By using the extracted features from a constrained dataset, we have developed a model that exceeds the detection accuracy of existing methods under similar limitations.

We then explore the multi-class classification of satellite antennas on ground terminals under various weather conditions. Specifically, we analyze conditions in two scenarios: the initial scenario where we classify antennas as either snow-covered or normal conditions, and the extended scenario that classifies antennas under snow, wet, and normal conditions. \hl{In the initial scenario with 80 training images, our proposed model achieves an accuracy of 88.33\% within 50 epochs, outperforming mainstream DL models including the top-performing R-YOLO and Faster R-CNN \mbox{\cite{NIPS2015_14bfa6bb}} models, whose accuracy values are 74.16\% and 80.00\%, after 500 epochs. In the extended scenario with 180 training images, our model reaches an accuracy of 88.33\% in 50 epochs, compared to R-YOLO and Faster R-CNN, whose accuracy is 72.22\% and 81.11\% after 500 epochs, respectively.}

While segmentation of satellite antennas using traditional YOLACT techniques is an initial step in our method, this paper focuses on the classification processes. Since the segmentation step follows standard YOLACT procedures, we have chosen to concentrate on the novel classification tasks in our study.

\section{System Model}
\subsection{Overview}

Our method utilizes feature-based TL to leverage the knowledge of a model trained on a weather database (source domain, \(D_S\)) and apply it to the satellite antenna classification task (target domain, \(D_T\)). We first consider the neural network as a function \(F(I; W)\) with input \(I\) and weights \(W\). The function \(F\) is initially trained on \(D_S\) to minimize the overall loss on \(D_S\), given by \(\sum_{(input, label) \in D_S} L(F(input, W), label)\), where \(D_S\) is the source domain, \(input\) is the input image from \(D_S\), \(label\) is the corresponding label, and \(L\) is the loss function of \(F\).

The goal of feature-based TL is to learn a weight \(OPT_T\) that minimizes the loss \(L(D_T, OPT_T)\). Given the optimal weight for the source domain \(OPT_S\), we initialize \(W = OPT_S\) and train \(W\) with data from the target domain \(D_T\) to minimize the loss \(L(D_T, W)\). Let \(MMD(D_S, D_T)\) denote the Maximum Mean Discrepancy \cite{zhuang2020comprehensive} between the characteristic distributions of \(D_S\) and \(D_T\). As \(MMD(D_S, D_T)\) decreases, the model's ability to generalize from \(D_S\) to \(D_T\) increases. 

\subsection{Pre-Processing: Background Removal Using YOLACT}

Background can introduce unnecessary complexity in \(D_T\), increasing the risk of overfitting and degrading model performance, as discussed by \cite{9156477}. To address this, we have fine-tuned YOLACT \cite{bolya2019yolact} specifically for the detection and segmentation of satellite dish antennas (collectively referred to as satellite dishes). We use YOLACT to first apply segmentation masks to the entire image, then filter out the Object of Interest \(O_I\) using the mask.

The application of YOLACT can be represented as a function \(Y\), transforming \(D_T\) into a simplified domain \(D_{ST}\), therefore focusing on \(O_I\) and aiming to minimize \(MMD\) between \(D_S\) and \(D_{ST}\). This step can be represented as \(Y: D_T \rightarrow D_{ST}\), and \(MMD(D_S, D_T) \geq MMD(D_S, D_{ST})\). YOLACT uses three primary losses to train the model: classification loss (\(L_{cls}\)), box regression loss (\(L_{box}\)), and mask loss (\(L_{mask}\)). The weights for these losses are 1, 1.5, and 6.125, respectively.

\begin{equation}
\label{eq_YOLACT}
L_{YOLACT} = L_{cls} + 1.5 \cdot L_{box} + 6.125 \cdot L_{mask}
\end{equation}

\begin{figure}[h]
    \centering
    \includegraphics[width=1\linewidth]{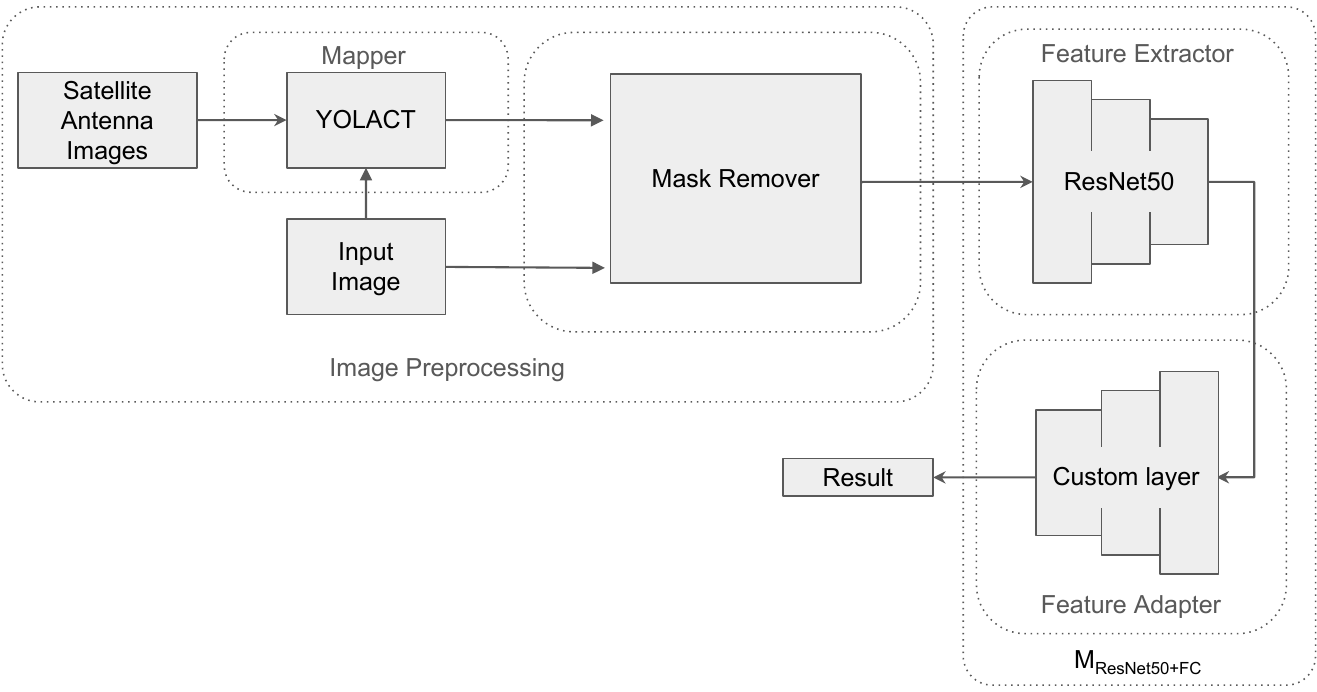}
    \caption{The process and architecture of the proposed transfer learning method.
    }
    \label{fig:system-diagram}
\end{figure}\vspace{-15pt}

\subsection{Proposed TL Method}
Our proposed method is shown in Fig. \ref{fig:system-diagram}, beginning with fine-tuning YOLACT using satellite antenna images to generate binary masks of satellite dishes in the images. After the fine-tuning stage, YOLACT will take input images and generate a mask. Masks and input images are then processed through Mask Remover to isolate satellite dishes in the image. The processed images are passed through a feature extractor, which extracts relevant features. These extracted features are used by the feature adapter to classify satellite dishes based on weather conditions. The higher layers of ResNet50 are task-specific and should be adjusted with a fine-tuned learning rate to effectively specialize in our task. We have used partial freezing and learning rate adjustment to preserve the generic features learning from \(D_S\) for the pre-trained model while allowing the model to adapt to the \(D_{ST}\).

Let \(W = {W_1, W_2, ..., W_n}\) represent the set of weights in the model, where \(W_n\) represent the weight of the last layer. The updates to \(W_i\) during the training process on \(D_{ST}\) can be represented by \(\delta W_i = \alpha \times \nabla L(D_{ST}; W) \) where \(\alpha\) is the learning rate. During the fine-tuning process, we have set the \(\alpha\) for \(W_1, W_2, ..., W_{n-1}\) to zero, and downscale \(\alpha\) to a lower value \(\alpha'\), So 
$
\delta W_i = \begin{cases}
    \alpha' \times \nabla L(D_{ST}; W),& \text{for } i = n\\
    0,              & \text{otherwise}
\end{cases}
$
\subsection{Model Architecture}
Our model, referred to as \( \text{M}_{\text{ResNet50+FC}}\), is based on pre-trained ResNet50. The last layer of ResNet50 is removed to use the model as a feature extractor, and the output features from the convolutions base are fed in the custom Fully Connected (FC) layers. The choice of ResNet50 is pivotal due to its deep residual learning framework, as \cite{He2015Deep} demonstrates, addressing the problem of the vanishing gradient. The study \cite{Al-Haija2022Detection} also demonstrates the effectiveness of ResNet50 in categorizing weather conditions. We have added two FC layers to learn the features extracted from ResNet50. The first FC layer contains 128 units followed by a dropout layer to prevent overfitting. The second FC layer matches the number of classes in the multi-class classification tasks.

Our model's loss function is cross-entropy loss, defined as 
\begin{equation}
\label{eq_cross_entropy}
    L(y, \hat{y}) = -[ylog(\hat{y}) + (1-y)log(1-\hat{y})]
\end{equation}

After combining the loss in the procedures of the TL, the final loss is given as 
\begin{equation}
\label{eq_lost}
    L_{overall} = L(y, \hat{y}) + L_{YOLACT},
\end{equation}
where $L_{YOLACT}$ is the segmentation loss and $L(y, \hat{y})$ is the classification loss.

\begin{figure}
    \centering
    \includegraphics[width=1\linewidth]{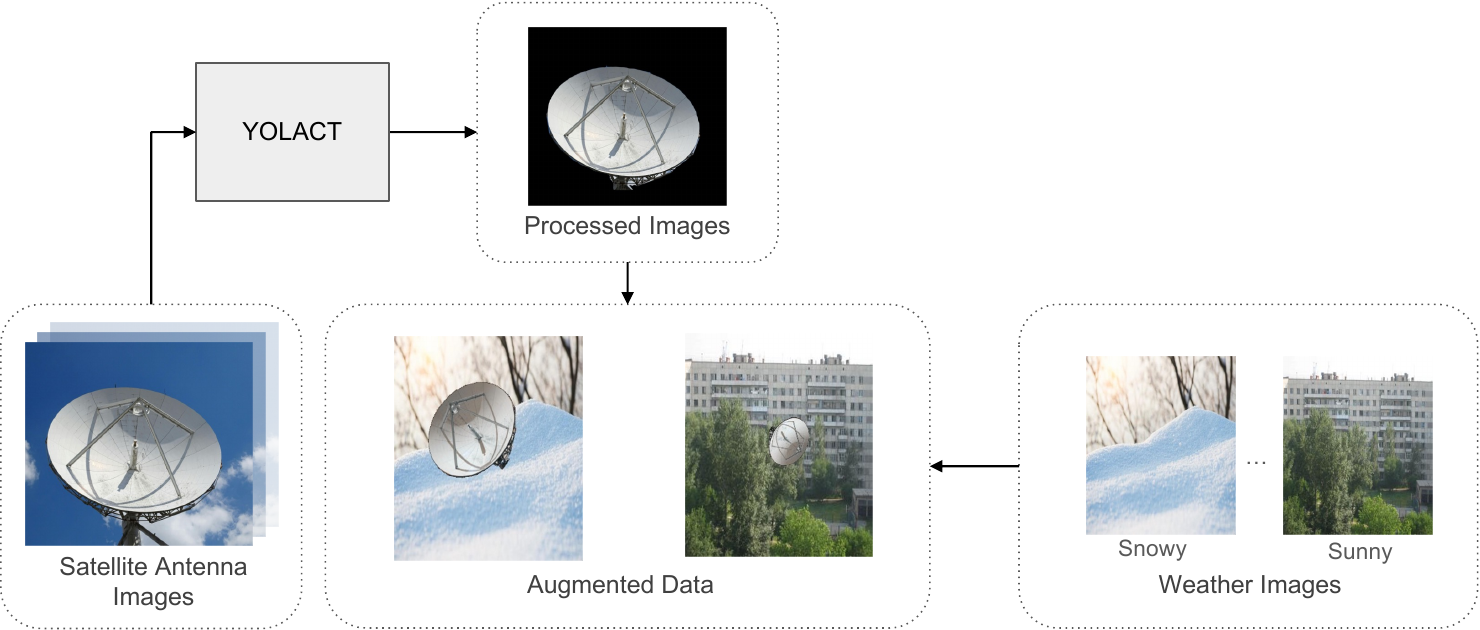}
    \caption{A process of data generation. Satellite dishes from original images are segmented using YOLACT and combined with various weather condition backgrounds to generate a diverse training images.}
    \label{fig:data_generation_process}
    
\end{figure}

\section{Data Preparation}
As shown in Fig. \ref{fig:data_generation_process}, we pre-processed our data to enhance the model's capability to identify satellite dishes in various weather conditions. We start with initial data preparation, where we collect images of satellite dishes under various conditions. To augment this dataset and enhance its diversity, we utilize two external datasets: the Weather Analysis Dataset \cite{ajayi2018multi} 
which provides images under typical weather conditions such as sunny and cloudy, and the Snow Pile Classification Dataset \cite{snow-pile-classification3_dataset}
which supplies images under snow conditions. For the extended scenario which includes wet conditions, we incorporate an additional dataset \cite{Qian_2018_CVPR} that offers images in rain conditions, \hl{ and we use DALL-E to generate images of satellite dishes in wet conditions due to the scarcity of available data.}

We then move to background removal, employing YOLACT to create binary masks. In this process, pixels corresponding to the object are assigned one binary value, distinguishing them as the $O_I$. After that, we use the binary mask to isolate the satellite dishes from their environments.

Finally, we perform image enhancement on our satellite dish dataset, which we have divided into two subsets: one for training and one for validation. Each subset is populated equally with images of satellite dishes captured under various weather conditions. For the training subset, we augment the satellite dishes by scaling, rotating, and merging them with background images from different weather scenarios to create a diverse set of training images. The same augmentation techniques are applied to the validation subset. In both subsets, the labels are associated with the satellite dishes. By doing so, we aim to test extreme scenarios where the conditions of the backgrounds and the satellite dishes differ significantly. Additionally, by separating satellite dishes in each subset, we prevent overfitting by ensuring that the model does not rely solely on the physical shapes of the satellite dishes for classification.

\section{Evaluation}

In this section, we discuss the experiment setup and evaluation metrics, followed by the evaluation results and results discussion.

\subsection{Evaluation Metrics}

Our model employs YOLACT, which focuses on segmentation and classification, in contrast to YOLO\hl{-based models}, which primarily focuses on object detection and classification. For this study, we focus exclusively on the classification loss component across all models.

Our evaluation strategy aligns with the model's focus on classification. We refine our evaluation to consider the following evaluation metrics: $AP = \frac{TP}{TP + FP}$, $mAP = \frac{1}{C} \sum_{i \in C} AP_i$, \hl{\mbox{$Accuracy = \frac{N_{correct}}{N}$}}, where \(TP\) is true positives, \(FP\) is false positives, \hl{\mbox{\(N\)} is the number of images,\mbox{~\(N_{correct}\)} is the number of images that get the correct prediction, and \mbox{\(C\)} is the number of objects classes.}

 \hl{The YOLO-based models utilize a composite loss function, including classification through cross-entropy, scaled by a factor \mbox{$\alpha = \frac{\text{C}}{80} \times \frac{3}{\text{nl}}$} which depends on the \mbox{\textbf{\(C\)}} and the number of layers ($nl$)}.

For our evaluation, we normalize the classification losses from \hl{the YOLO-based models} by dividing them by $\alpha$, adjusting them to a comparable scale. Similarly, for our model, we focus on the classification loss $L(y, \hat{y})$, to ensure consistency in the evaluation criteria between all models.

\subsection{Experimental Setup and Implementation Details}

Here we evaluate the proposed method in comparison with the \hl{YOLOv7 \mbox{\cite{wang2023yolov7}}, YOLOv9 \mbox{\cite{Wang2024YOLOv9LW}}, R-YOLO \mbox{\cite{9989413}}, and Faster R-CNN models, conducting experiments in Google Colab on an NVIDIA A100 GPU}. We did not apply parameter optimization techniques, such as genetic algorithms, as we want to maintain consistent configurations across all models and avoid extensive parameter tuning to ensure comparability.

\hl{We used the Faster R-CNN \mbox{\cite{NIPS2015_14bfa6bb}} model from the Detectron2 library and configured it with ResNet50 and Feature Pyramid Networks (FPN). Our experiments utilized a batch size of 8 and ran for 11,000 iterations, which is approximately equivalent to 500 epochs. Learning rate adjustments were applied at 70\% and 90\% of the total iterations, following a step-wise learning rate schedule. Other settings, including anchor sizes, aspect ratios, and region proposal parameters, remained at their default values.}

\hl{For comparison with the latest YOLO-based model, YOLOv9, we used the YOLOv9-m model (referred to as ``YOLOv9'' for short). In addition, we introduced a YOLOv9 variation called YOLOv9-Freeze, in which we applied the partial freezing technique to freeze the backbone of the YOLOv9-m model. The backbone of YOLOv9 is responsible for feature extraction, and by freezing it, we retain its pre-trained capabilities while allowing the head, which handles object classification, to be fine-tuned for the specific task. As partial freezing is used in our proposed model to enhance performance with limited data, YOLOv9-Freeze can be considered directly comparable to the proposed model.}

Our study focuses on multi-class classification tasks, evaluating our model on satellite dishes under various weather conditions, and all images used in our analysis are 300$\times$300 px in size. We first analyzed scenarios distinguishing between snow-covered and normal conditions, \hl{which is essentially a binary classification task}. We then extended our analysis to include wet conditions, broadening our model to a more comprehensive multi-class classification. 

For \hl{the classification tasks}, we used model \( \text{M}_{\text{ResNet50+FC}}\) and fine-tuned it on the weather dataset to distinguish between different weather conditions. The training parameters for \( \text{M}_{\text{ResNet50+FC}}\) include the Adam optimizer with an initial learning rate of $2 \times 10^{-4}$, and weight decay of $5 \times 10^{-4}$ to optimize the model performance.

To pre-process satellite dish images, we employed YOLACT and used the Adam optimizer with a learning rate of $2 \times 10^{-3}$ over 30,000 epochs to fine-tune its weights, ensuring the model focuses on the nuanced features of satellite dishes.

\subsection{Experimental Results}
The outcomes of our experiments highlight the classification capabilities of the proposed model compared to \hl{YOLOv7, YOLOv9, YOLOv9-Freeze, R-YOLO, and Faster R-CNN}. We analyzed the impact of the volume of training images on model performance through the evaluation process.

\hl{Based on the data preparation discussed in Section III, we constructed three distinct datasets for each scenario. In the initial scenario, each dataset was paired with a constant test set of 120 images. In the extended scenario, we utilized a larger test set of 180 images to provide a more comprehensive evaluation. Tables \mbox{\ref{table_binary_classification}-\ref{table_multiclass_classification} }summarize the performance of each model in comparison to the baseline methods.}

\hl{We used four background conditions: snow, sunny, cloudy, and rain. To maintain consistency, each dataset contained an equal number of images for every combination of background conditions and satellite dish conditions. In the initial scenario, this resulted in eight combinations by pairing the four background conditions with two types of satellite dishes. For the dataset with 40 images, we generated five images for each combination. For the dataset with 64 images, we generated eight images per combination. For the dataset with 80 images, we generated 10 images per combination.}

\hl{In the extended scenario, we added the wet condition satellite dish, leading to 12 combinations. For the dataset containing 60 images, we generated five images for each combination. For the dataset with 120 images, we generated 10 images per combination. For the dataset with 180 images, we generated 15 images per combination.}

We compared our proposed model with YOLOv7\hl{, YOLOv9, YOLOv9-Freeze, R-YOLO, and Faster R-CNN} using the datasets we created. The evaluation metrics for each result were recorded upon completion of the training cycles.

\begin{table}[ht]
\setlength{\tabcolsep}{0.5em} 
\renewcommand{\arraystretch}{1}
\caption{\hl{Performance Comparison in the Initial Scenario: Snow and Normal Conditions}}
\label{table_binary_classification}
\centering
\begin{tabular}{cccc}
\toprule
{\bfseries Model} & {\bfseries Training Images} & {\bfseries mAP} & {\bfseries Accuracy} \\
\midrule
YOLOv7 & 40 & 24.78\% & 49.16\% \\
YOLOv9 & 40 & 45.55\% & 46.66\% \\
YOLOv9-Freeze & 40 & 69.25\% & 61.66\% \\
R-YOLO & 40 & 64.51\% & 63.33\% \\
Faster R-CNN & 40 & 75.84\% &  55.83\% \\
Proposed Model & 40 & {\bfseries77.69\%} & {\bfseries78.33\%} \\
\midrule
YOLOv7 & 64 & 47.00\% & 48.33\% \\
YOLOv9 & 64 & 52.51\% & 52.50\% \\
YOLOv9-Freeze & 64 & 69.02\% & 67.50\% \\
R-YOLO & 64 & 69.14\% & 68.16\% \\
Faster R-CNN & 64 & 67.77\% &  63.33\% \\
Proposed Model & 64 & {\bfseries82.13\%} & {\bfseries85.00\%} \\
\midrule
YOLOv7 & 80 & 58.77\% & 51.66\% \\
YOLOv9 & 80 & 64.44\% & 53.33\% \\
YOLOv9-Freeze & 80 & 65.55\% & 66.66\% \\
R-YOLO & 80 & 74.22\% & 74.16\% \\
Faster R-CNN & 80 & 82.29\% & 80.00\% \\
Proposed Model & 80 & {\bfseries87.28\%} & {\bfseries88.33\%} \\
\bottomrule
\end{tabular}
\end{table}

\begin{table}[ht]
\setlength{\tabcolsep}{0.5em} 
\renewcommand{\arraystretch}{1}
\caption{\hl{Performance Comparison in the Extended Scenario: Snow, Wet, and Normal Conditions}}
\label{table_multiclass_classification}
\centering
\begin{tabular}{cccc}
\toprule
{\bfseries Model} & {\bfseries Training Images} & {\bfseries mAP}& {\bfseries Accuracy} \\
\midrule
YOLOv7 & 60 & 18.90\% & 33.88\% \\
YOLOv9 & 60 & 37.27\% & 34.44\% \\
YOLOv9-Freeze & 60 & 45.00\% & 43.88\% \\
R-YOLO & 60 & 53.86\% & 49.44\% \\
Faster R-CNN& 60 & 58.89\% & 59.44\% \\
Proposed Model & 60 & {\bfseries67.25\%} & {\bfseries66.66\%} \\
\midrule
YOLOv7 & 120 & 33.88\% & 36.66\% \\
YOLOv9 & 120 & 35.12\% & 37.22\% \\
YOLOv9-Freeze & 120 & 51.88\% & 52.22\% \\
R-YOLO & 120 & 57.58\% & 60.55\% \\
Faster R-CNN& 120 & 72.89\% & 72.77\% \\
Proposed Model & 120 & {\bfseries75.26\%} & {\bfseries77.22\%} \\
\midrule
YOLOv7 & 180 & 23.95\% & 35.55\% \\
YOLOv9 & 180 & 37.27\% & 40.55\% \\
YOLOv9-Freeze & 180 & 61.52\% & 55.55\% \\
R-YOLO & 180 & 71.57\% & 72.22\% \\
Faster R-CNN& 180 & 79.75\% & 81.11\% \\
Proposed Model & 180 & {\bfseries 86.46\%} & {\bfseries88.33\%} \\
\bottomrule
\end{tabular}
\end{table}

\begin{table}[ht]
\setlength{\tabcolsep}{0.5em} 
\renewcommand{\arraystretch}{1}
\caption{\hl{Performance Comparison with Satellite Antenna Images: Snow, Wet, and Normal Conditions }}
\label{table_real}
\centering
\begin{tabular}{ccccc}
\toprule
{\bfseries Model} & {\bfseries mAP} & {\bfseries Accuracy} \\
\midrule
YOLOv7  & 23.95\% & 35.40\% \\
YOLOv9  & 37.83\% & 44.44\% \\
YOLOv9-Freeze & 50.99\% & 46.51\% \\
R-YOLO  & 72.27\% & 62.22\% \\
Faster R-CNN & 78.33\% & 73.33\% \\
Proposed Model & {\bfseries 82.18\%} & {\bfseries86.67\%} \\
\bottomrule
\end{tabular}
\end{table}

\subsection{Initial Scenario}
Table~\ref{table_binary_classification} \hl{ shows that our model outperforms YOLOv7, YOLOv9, YOLOv9-Freeze, R-YOLO, and Faster R-CNN in terms of mAP and accuracy within the initial scenario, which distinguishes between two classes: snow and normal. With 40 training images, our model achieves an accuracy of 78.33\%, followed by R-YOLO's 63.33\% and YOLOv9-Freeze's 61.66\%. With 80 training images, our model reaches an accuracy of 88.33\%, compared to 80.00\% for Faster R-CNN and 74.16\% for R-YOLO.}

\subsection{Extended Scenario}
 
Table~\ref{table_multiclass_classification} \hl{compares the performance of models in multi-class classification, specifically among three classes: snow, wet, and normal. Our model consistently demonstrates superior performance across different sizes of training sets. With 60 training images, our proposed model achieves an accuracy of 66.66\%, followed by Faster R-CNN's 59.44\% and R-YOLO's 49.44\%. As the training dataset expands to 180 images, our model reaches an accuracy of 88.33\%, while Faster R-CNN remain in second place at 81.11\%, and R-YOLO comes in third at 72.22\%.}

\subsection{Discussion of Model Superiority}
\hl{Table\mbox{~\ref{table_binary_classification}} and Table\mbox{~\ref{table_multiclass_classification} }show that our proposed model consistently outperforms R-YOLO, Faster R-CNN, and other state-of-the-art models.} Our application of the proposed TL method divides the overall loss function in (\ref{eq_lost}) into two components: \(L_{YOLACT}\) and \(L^*\). The segmentation component, represented by \(L_{YOLACT}\), isolates the target object from its background, allowing for more focused classification. The other component, \(L^*\), represents the loss of a customized core classification for our specific task. Due to the reduced \(MMD\), \(L^*\) is already closer to the optimal configuration for our targeted classification task. Therefore, fine-tuning \(L^*\) becomes straightforward, focusing on minor adjustments to adapt the model to the target task, which reduces the need for large training data.

\hl{In comparison, single-shot models such as YOLOv7 and YOLOv9 show lower performance due to their reliance on large and diverse datasets, as discussed by Zhang \textit{et al.}\mbox{ ~\cite{zhang2017singleshot}}. YOLOv9-Freeze achieves higher accuracy than YOLOv9, as it benefits from its pre-trained backbone, which utilizes knowledge gained from the COCO dataset to enhance feature extraction capabilities. This use of TL enables YOLOv9-Freeze to require fewer training images, resulting in higher accuracy. R-YOLO benefits from FCNet's feature calibration modules by effectively aligning features across weather conditions, decreasing the need for large and diverse datasets. Faster R-CNN shows strong results due to its two-stage classification process, which trades off processing time for improved performance.}

\subsection{Generalizability to Real-World Scenarios}
\hl{We trained the models using 180 augmented satellite antenna images from the extended scenario and tested the them on 45 unaugmented satellite antenna images, 15 per weather condition. As shown in Table\mbox{~\ref{table_real}}, our proposed model achieved the highest accuracy of 86.67\%, followed by Faster R-CNN at 78.33\% and R-YOLO at 62.22\%. The accuracy of our model decreased by about 1.67\% compared to the results for the extended scenario, while the accuracy for Faster R-CNN and R-YOLO dropped by approximately 7.78\% and 11.94\%. This smaller decrease in accuracy demonstrates that our model effectively generalizes to real-world conditions compared to Faster R-CNN and R-YOLO. YOLOv7 and YOLOv7-Freeze maintained low accuracy, conversely, YOLOv9's slight increase in accuracy suggests that it may perform better with this simpler, unaugmented data.}

\begin{figure}
    \centering
    \includegraphics[width=1\linewidth]{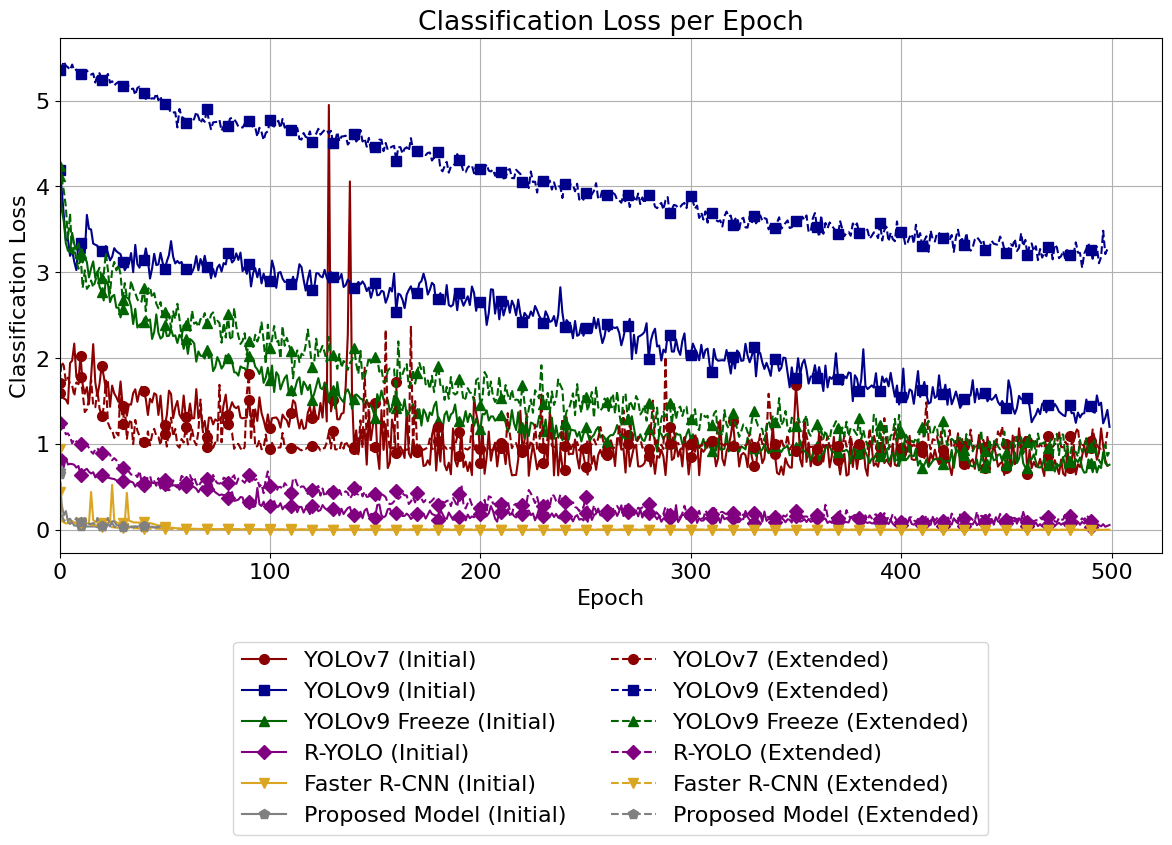}
    \caption{\hl{Loss curve comparison across epochs for the proposed model and other models. The results are grouped by the ``Initial'' and ``Extended'' scenarios.} }
    \label{fig:loss}
\end{figure}

\subsection{Analysis of Loss Curves}

Loss curves show the efficiency of a model's learning by tracking the reduction in error across training epochs. Fig. \ref{fig:loss} compares the classification \hl{losses} between epochs for our model, YOLOv7, \hl{YOLOv9, YOLOv9-Freeze, R-YOLO, and Faster R-CNN in both the initial and extended scenarios.}

In the initial scenario, our model achieves a classification loss of 0.03 within \hl{50 epochs. In contrast, YOLOv7 and R-YOLO reach minimum losses of 0.62 and 0.03 after 500 epochs, respectively. YOLOv9 reaches a loss of 1.20, while YOLOv9-Freeze reaches a loss of 0.666. Faster R-CNN achieves a low loss of 0.0009. In the extended scenario, our model again achieves a loss of 0.03 within 50 epochs. In comparison, YOLOv7 and R-YOLO reach minimum losses of 0.90 and 0.134 after 500 epochs. YOLOv9 and YOLOv9-Freeze show losses of 3.05 and 0.76, while Faster R-CNN achieves a loss of 0.0011.}

\hl{In the extended scenario, our proposed model achieves an accuracy of 88.33\%, which is higher than the 80.00\% accuracy of Faster R-CNN. However, Faster R-CNN achieves a lower classification loss of 0.001, while our model shows a loss of 0.03. The lower classification loss observed in Faster R-CNN suggests more efficient fitting to the training data. However, it does not fully capture the features required to distinguish satellite dishes. In contrast, our model utilizes TL by incorporating a pre-trained ResNet50, initially trained on simpler weather detection tasks, alongside YOLACT for segmentation. This allows our model to leverage generalized weather detection knowledge and apply it to this specialized classification task. As a result, our model achieves higher accuracy, demonstrating the advantage of TL in adapting to real-world variations.}

\hl{A similar pattern is observed when comparing our proposed model to R-YOLO in the initial scenario. Our model delivers higher accuracy at 88.33\%, compared to R-YOLO's 74.16\%, while maintaining a similar classification loss of 0.03.}

\begin{table}[ht]
\setlength{\tabcolsep}{0.5em} 
\renewcommand{\arraystretch}{1.2} 
\caption{\hl{Computational Complexity Comparison}}
\label{table_computational_complexity}
\centering
\begin{tabular}{lrr}
\toprule
{\bfseries Model} & {\bfseries GFLOPs } & {\bfseries Memory (GB)} \\
\midrule
Proposed Model & 186.0 & 16.3 \\
R-YOLO & 51.2 & 14.2 \\
YOLOv7 & 105.2 & 4.2 \\
YOLOv9 & 132.4 & 28.8 \\
YOLOv9-Freeze & 132.4 & 28.8 \\
Faster R-CNN & 238.8 & 18.5 \\
\bottomrule
\end{tabular}
\end{table}\vspace{-18pt}

\subsection{Computational Complexity Analysis}
\hl{During the model training phase, we recorded the Giga Floating-Point Operations (GFLOPs) and memory usage. To keep the analysis consistent, we use the input size of 640 \mbox{\(\times\)} 640 px for all models, which is the same as the default input size for YOLO-based models and not the default input size of Faster R-CNN model configured on Detectron2 (as Detectron2 uses dynamic input sizes by default). Memory usage was recorded based on a batch size of 16, increasing the batch size would result in higher memory usage.}

\hl{Table \mbox{\ref{table_computational_complexity}} shows the computational complexity, particularly focusing on GFLOPs and memory consumption.}

\hl{The total GFLOPs value for our model is calculated as:}

\begin{equation}
\label{eq_GFLOPs}
\begin{split}
\text{GFLOPs}_{\text{total}} = \text{GFLOPs}_{\text{YOLACT}} + \text{GFLOPs}_{\text{remover}} +  \\ \text{GFLOPs}_{ \text{M}_{\text{ResNet50+FC}}}
\end{split}
\end{equation}

\hl{Where \mbox{\(\text{GFLOPs}_{\text{YOLACT}}\)} refers to the GFLOPs used by the YOLACT model, \mbox{\(\text{GFLOPs}_{\text{remover}}\)} refers to the GFLOPs consumed by the mask remover, and \mbox{\(\text{GFLOPs}_{\text{M}_{\text{ResNet50+FC}}}\)} refers to the GFLOPs used by the prediction model.}

\hl{The GFLOPs value of our model is calculated as the sum of the GFLOPs consumed by YOLACT, the mask remover, and \mbox{\(\text{M}_{\text{ResNet50+FC}}\)}. For our specific implementation, YOLACT contributes significantly to the total GFLOPs of 118.6, making it the primary factor in the model's computational complexity. The \mbox{\(\text{M}_{\text{ResNet50+FC}}\)} component requires 67.4 GFLOPs, while the mask remover adds an insignificant amount of GFLOPs because its operations are straightforward and scale linearly with the image size. In total, this amounts to approximately 186.0 GFLOPs. In comparison, R-YOLO requires 51.2 GFLOPs, YOLOv7 requires 105.2 GFLOPs, YOLOv9 requires 132.4 GFLOPs, and Faster R-CNN requires 238.8 GFLOPs.}

\hl{In terms of memory consumption, the memory usage during the training phase is determined by the most memory-intensive component in the series of operations. The memory usage for our model is calculated as:}

\begin{equation}
\label{eq_memory}
\begin{split}
\text{Mem}_{\text{total}} = \max(\text{Mem}_{\text{YOLACT}}, \text{Mem}_{\text{remover}}, 
\text{Mem}_{\text{M}_{\text{ResNet50+FC}}})
\end{split}
\end{equation}

\hl{In (5), \mbox{\(\text{Mem}_{\text{YOLACT}}\)} refers to the memory used by the YOLACT model, \mbox{\(\text{Mem}_{\text{remover}}\)} denotes the memory consumed by the mask remover, and \mbox{\(\text{Mem}_{\text{M}_{\text{ResNet50+FC}}}\)} represents the memory used by the prediction model. Our model requires 16.3 GB of memory, which is higher than R-YOLO's 14.2 GB and YOLOv7's 4.2 GB but lower than YOLOv9's 28.8 GB and Faster R-CNN's 18.5 GB.}

\hl{While YOLACT provides solid segmentation capabilities, its heavy computational and memory demands present a trade-off. This trade-off is reflected in our model's accuracy and mAP metrics. There is potential for optimizing the model by replacing YOLACT with a simpler image segmentation tool. Such a substitution could reduce both the GFLOPs and memory usage. This possibility can be explored in future work, where the trade-offs between segmentation quality and computational efficiency are carefully balanced.}

\subsection{Discussion of Deployment Options}
A general commercial off-the-shelf camera that can generate at least 300$\times$300 px images (which is the image size we used in the paper) with its field of view covering the ground terminal being monitored will suffice. \hl{To install, the camera should be mounted in a fixed position, ensuring its view consistently captures the entire satellite antenna. Calibration may be needed to adjust angles and avoid interference from environmental factors such as lighting or obstacles. Once mounted and calibrated, the camera can operate continuously or on a scheduled basis.} Although we already discussed the scenarios with representative weather conditions, our proposed method with image pre-processing and $\text{M}_{\text{ResNet50+FC}}$ model can be extended to work in various forms factors of satellite antennas (e.g., square phased array antennas, large parabolic antennas on Earth stations etc.) and scenarios with additional weather conditions. Due to the generalization of the method, it can be also applied to telescopes used in an optical ground station, which is expected to be used in future satellite Internet.

\hl{It is worth noting that in real-world deployments the system may encounter uniquely designed antennas that differ significantly from the samples used to train the YOLACT models. In such cases, the system might struggle to classify the weather conditions on these antennas accurately, potentially reducing detection accuracy. A solution to this would be expanding the training dataset to include a broader variety of antenna designs.}

\hl{While we utilized an NVIDIA A100 GPU for training and testing, the proposed method can still operate on lower-end GPUs. A less powerful GPU will primarily affect inference time, and if the system's memory capacity is limited, reducing the batch size during inference can help relax memory requirements.}

\section{Conclusion}
The accurate detection of fine-grained conditions on ground terminals resulting from adverse weather events has become a challenge to realizing reliable satellite Internet enabled by the modern LEO satellite networks. The proposed TL-based method provides a novel approach to addressing the challenge. Based on the evaluation results, our model can learn and generalize effectively from a minimal number of training images, outperforming \hl{YOLOv7, YOLOv9, YOLOv9-Freeze, R-YOLO, and Faster R-CNN} under conditions of limited data. This highlights our method's potential for object classification applications where data availability is constrained. \hl{In practical deployment, the model's ability to achieve high performance with minimal training data reduces the need for extensive data collection and training, making the method particularly suitable for scenarios where acquiring large datasets is infeasible. With limited training data, it can handle diverse weather conditions and ensure reliable performance in real-world satellite ground terminals.} The proposed TL method can be deployed standalone or as part of the satellite antennas and effortlessly extended to ground terminal components whose performance and operations are subject to effects caused by weather conditions.



\normalsize

\bibliographystyle{IEEEtran}
\bibliography{./references}

\begin{thebibliography}{10}
\providecommand{\url}[1]{#1}
\csname url@samestyle\endcsname
\providecommand{\newblock}{\relax}
\providecommand{\bibinfo}[2]{#2}
\providecommand{\BIBentrySTDinterwordspacing}{\spaceskip=0pt\relax}
\providecommand{\BIBentryALTinterwordstretchfactor}{4}
\providecommand{\BIBentryALTinterwordspacing}{\spaceskip=\fontdimen2\font plus
\BIBentryALTinterwordstretchfactor\fontdimen3\font minus \fontdimen4\font\relax}
\providecommand{\BIBforeignlanguage}[2]{{%
\expandafter\ifx\csname l@#1\endcsname\relax
\typeout{** WARNING: IEEEtran.bst: No hyphenation pattern has been}%
\typeout{** loaded for the language `#1'. Using the pattern for}%
\typeout{** the default language instead.}%
\else
\language=\csname l@#1\endcsname
\fi
#2}}
\providecommand{\BIBdecl}{\relax}
\BIBdecl

\bibitem{Padmanabhan19}
R.~Padmanabhan, A.~Schulman, D.~Levin \emph{et~al.}, ``Residential links under the weather,'' in \emph{Proc. of the ACM SIGCOM}.\hskip 1em plus 0.5em minus 0.4em\relax New York, NY, USA: ACM, 2019, p. 145–158.

\bibitem{Laniewski2024WetLinksAL}
D.~Laniewski, E.~Lanfer, B.~Meijerink \emph{et~al.}, ``Wetlinks: a large-scale longitudinal starlink dataset with contiguous weather data,'' \emph{arXiv:2402.16448 [cs.NI]}, 2024.

\bibitem{girshick2014rich}
R.~Girshick, J.~Donahue, T.~Darrell \emph{et~al.}, ``Rich feature hierarchies for accurate object detection and semantic segmentation,'' in \emph{2014 IEEE CVPR}, 2014, pp. 580--587.

\bibitem{wang2023yolov7}
C.-Y. Wang, A.~Bochkovskiy, and H.-Y.~M. Liao, ``Yolov7: Trainable bag-of-freebies sets new state-of-the-art for real-time object detectors,'' in \emph{2023 IEEE/CVF CVPR}, 2023, pp. 7464--7475.

\bibitem{10.1007/978-3-319-46448-0_2}
W.~Liu, D.~Anguelov, D.~Erhan \emph{et~al.}, ``Ssd: Single shot multibox detector,'' in \emph{Computer Vision -- ECCV 2016}, B.~Leibe, J.~Matas, N.~Sebe \emph{et~al.}, Eds.\hskip 1em plus 0.5em minus 0.4em\relax Cham: Springer International Publishing, 2016, pp. 21--37.

\bibitem{Leng2019An}
J.~Leng and Y.~Liu, ``An enhanced ssd with feature fusion and visual reasoning for object detection,'' \emph{Neural Computing and Applications}, pp. 1--10, 2019.

\bibitem{9989413}
L.~Wang, H.~Qin, X.~Zhou \emph{et~al.}, ``R-yolo: A robust object detector in adverse weather,'' \emph{IEEE Transactions on Instrumentation and Measurement}, vol.~72, pp. 1--11, 2023.

\bibitem{Xu2020Transfer}
W.~Xu, J.~He, and Y.~Shu, ``Transfer learning and deep domain adaptation,'' in \emph{Advances and Applications in Deep Learning}, M.~A. Aceves-Fernandez, Ed.\hskip 1em plus 0.5em minus 0.4em\relax Rijeka: IntechOpen, 2020, ch.~3.

\bibitem{NIPS2015_14bfa6bb}
S.~Ren, K.~He, R.~Girshick \emph{et~al.}, ``Faster r-cnn: Towards real-time object detection with region proposal networks,'' in \emph{Advances in Neural Information Processing Systems}, vol.~28.\hskip 1em plus 0.5em minus 0.4em\relax Curran Associates, Inc., 2015.

\bibitem{zhuang2020comprehensive}
F.~Zhuang, Z.~Qi, K.~Duan \emph{et~al.}, ``A comprehensive survey on transfer learning,'' \emph{Proceedings of the IEEE}, vol. 109, no.~1, pp. 43--76, 2021.

\bibitem{9156477}
Y.~Shen, R.~Ji, Z.~Chen \emph{et~al.}, ``Noise-aware fully webly supervised object detection,'' in \emph{2020 IEEE/CVF Conference on Computer Vision and Pattern Recognition (CVPR)}, 2020, pp. 11\,323--11\,332.

\bibitem{bolya2019yolact}
D.~Bolya, C.~Zhou, F.~Xiao \emph{et~al.}, ``Yolact: Real-time instance segmentation,'' in \emph{2019 IEEE/CVF International Conference on Computer Vision (ICCV)}.\hskip 1em plus 0.5em minus 0.4em\relax Los Alamitos, CA, USA: IEEE Computer Society, nov 2019, pp. 9156--9165.

\bibitem{He2015Deep}
K.~He, X.~Zhang, S.~Ren \emph{et~al.}, ``Deep residual learning for image recognition,'' \emph{2016 IEEE CVPR}, pp. 770--778, 2015.

\bibitem{Al-Haija2022Detection}
Q.~A. Al-Haija, M.~Gharaibeh, and A.~Odeh, ``Detection in adverse weather conditions for autonomous vehicles via deep learning,'' \emph{AI}, 2022.

\bibitem{ajayi2018multi}
\BIBentryALTinterwordspacing
G.~Ajayi, ``Multi-class weather dataset for image classification,'' 2018, accessed on 2024-03-15. [Online]. Available: \url{https://doi.org/10.17632/4drtyfjtfy.1}
\BIBentrySTDinterwordspacing

\bibitem{snow-pile-classification3_dataset}
\BIBentryALTinterwordspacing
U.~of~Minnesota, ``Snow pile classification3 dataset,'' jun 2023. [Online]. Available: \url{https://universe.roboflow.com/university-of-minnesota-cmnhh/snow-pile-classification3}
\BIBentrySTDinterwordspacing

\bibitem{Qian_2018_CVPR}
R.~Qian, R.~T. Tan, W.~Yang \emph{et~al.}, ``Attentive generative adversarial network for raindrop removal from a single image,'' in \emph{The IEEE Conference on Computer Vision and Pattern Recognition (CVPR)}, June 2018.

\bibitem{Wang2024YOLOv9LW}
C.-Y. Wang, I.-H. Yeh, and H.~Liao, ``Yolov9: Learning what you want to learn using programmable gradient information,'' \emph{arXiv:2402.13616v2 [cs.CV]}, 2024.

\bibitem{zhang2017singleshot}
Z.~Zhang, S.~Qiao, C.~Xie \emph{et~al.}, ``Single-shot object detection with enriched semantics,'' \emph{2018 IEEE/CVF Conference on Computer Vision and Pattern Recognition}, pp. 5813--5821, 2017.

\end{thebibliography}

\vfill

\end{document}